\documentclass{extarticle}

\usepackage{arxiv}

\usepackage[utf8]{inputenc} 
\usepackage[T1]{fontenc}    
\usepackage{hyperref}       
\usepackage{url}            
\usepackage{booktabs}       
\usepackage{amsfonts}       
\usepackage{nicefrac}       
\usepackage{microtype}      
\usepackage{lipsum}
\usepackage{graphicx}
\usepackage{tabularx}
\graphicspath{ {./images/} }
\usepackage[font=small,width=0.9\textwidth]{caption}
\usepackage{comment}
\graphicspath{{Figures/}}
\usepackage{moresize}
\usepackage{amssymb}
\usepackage{amsmath}
\usepackage[section]{placeins}
\usepackage{caption}
\usepackage{float}

\title{A Computational Model of Learning and Memory
Using Structurally Dynamic Cellular Automata}

\author{
 Jeet Singh \\
 Boston, MA 02108\\
 \texttt{jeetsingh@alum.mit.edu} \\
}
\begin{document}
\maketitle
\begin{abstract}
In the fields of computation and neuroscience, much is still unknown about the underlying computations that enable key cognitive functions including learning, memory, abstraction and behavior. This paper proposes a mathematical and computational model of learning and memory based on a small set of bio-plausible functions that include coincidence detection, signal modulation, and reward/penalty mechanisms. Our theoretical approach proposes that these basic functions are sufficient to establish and modulate an information space over which computation can be carried out, generating signal gradients usable for inference and behavior.  The computational method used to test this is a structurally dynamic cellular automaton with continuous-valued cell states and a series of recursive steps propagating over an undirected graph with the memory function embedded entirely in the creation and modulation of graph edges. The experimental results show:  that the toy model can make near-optimal choices to re-discover a reward state after a single training run; that it can avoid complex penalty configurations; that signal modulation and network plasticity can generate exploratory behaviors in sparse reward environments; that the model generates context-dependent memory representations; and that it exhibits high computational efficiency because of its minimal, single-pass training requirements combined with  flexible and contextual memory representation.
\end{abstract}

\section{Introduction}
	Despite rapid advances in the field of cognitive neuroscience and increasing understanding of the underlying biological components, structures, locations and electrochemical activity in the brain --- that is, the substrate on which biological computation must be built, there is limited insight into what types of computations are being performed to allow such complex and energy-efficient capabilities in even extremely simple living organisms. Other key constructs that we rely on to define and support “cognition”, such as learning, memory, abstraction, behavior, motivation, exploration, and others are also not well understood or effectively modeled in a mathematical or axiomatic fashion.

  	In the related machine intelligence sub-field of artificial general intelligence (AGI) that addresses generalized machine learning, several techniques such as reinforcement learning\cite{sutton2018reinforcement}, often in combination with neural network and deep-learning methods, have been very successful in specific contexts such as game-playing at human or even super-human levels. However, the enormous amount of training data and real-time computational requirements of many of these models prove challenging in dynamic real-world scenarios where the overall environment is not fully specified or known to the artificial agent.  Additionally, some critical elements of AGI such as the ability to deal with compositionality, or combinatorial generalization\cite{battaglia2018relational}, as well as continual or lifetime learning\cite{kudithipudi2022biological,van2020brain} in multi-task environments remains difficult to achieve.

   What are the fundamental mathematical elements of learning that might span both natural and machine learning? This paper proposes a mathematical and computational model of learning, memory and behavioral functions that allows an agent to navigate an environment, respond to sensory cues, make “decisions” and/or take actions in response to positive signals\cite{silver2021reward} (“rewards”) or negative ones (“penalties”). Making near-optimal choices consistently and effectively is a common and not unreasonable definition of what we might call learning. In its design, the model addresses three broad questions posed below:
\begin{enumerate}
\item
If we assume that computation must be happening, then  how is the information space established in the first place, based on external sensory inputs, and how is computation carried out over that space?
\item 
What type of axiomatic information representation and manipulation model could help address some of the limiting factors in current AI methods\cite{hassabis2017neuroscience}, such as the need for large amounts of sample or training data, high computational requirements, and difficulty in generalized learning and task-switching?
\item
Based on our current understanding of neuroscience, what are some bio-plausible functions that could underpin the prior two questions?
\end{enumerate}

To address the latter requirement of bio-plausibility, the model is restricted to using a very small set of basic functions that we suggest are likely to be available to neural circuitry: these include coincidence detection (the ability to recognize and store simultaneous sensory inputs in a pseudo-Hebbian fashion), signal modulation, and reward/penalty functions. These functions are all recognizably important in control systems\cite{todorov2009efficient}. We do not attempt to describe the specific neurobiological mechanisms that might produce these base functions, but do assume some potential capabilities and constraints to guide the development of the model.

The primary goal in constructing the model is to define a set of algorithms that, when reacting to a set of input stimuli can generate experience/behavior outputs that resemble basic cognitive phenomena, including constructs such as learning, memory, inference, and behavior. The key target behavior of the model is to generate optimal choices in an information “landscape”, which is expressed as an environment graph.

The use of a structurally dynamic cellular automaton (CA) as the underlying computational model allows for the inherent parallelism of CAs to be utilized as analogous to neural computation, with network plasticity being achieved by using both the cell update function as well as an update function that modifies the underlying network or undirected graph that the CA consists of. Although the model could also be specified in the terminology of a convolutional recurrent neural network, or other flavors of graph cellular automata\cite{grattarola2021learning} or even reservoir computing (RC), the choice of cellular automata provides a simple and concise way to express this novel computational model, which we call Coincident Graph Learning (CGL).

A toy model is specified mathematically and algorithmically (in MATLAB), and tested in small simulated environments of graph networks of varying complexity - from highly ordered lattice-like networks approximating reinforcement learning grid world tasks to a range of less and more randomized Watts-Strogatz small world networks\cite{watts1998collective}. A variety of reward and penalty scenarios are tested, as well as multi-reward and multi-penalty scenarios. All the simulations involve the agent being trained on an environment containing a reward (or penalty, or both), and then being placed back somewhere in the environment to navigate to the reward. In all cases, results are based on a single training run through the environment. 

\section{Theoretical Approach}
In most if not all machine learning algorithms and applications, a key distance metric must be defined that provides the basis for optimization/maximization/minimization to generate a useful solution. In other words, the information space needs to be established. This may take the form of the value equation in a reinforcement learning algorithm, or the surfaces being analyzed in a gradient descent approach in deep learning. But what sort of metric could be used in a biological setting? Establishing this underlying structure is a key problem to be addressed in a system that does not have machine-like data inputs, labels, or fixed geometries, whether Euclidean or non-Euclidean. The selection of a mechanism that could be used to generate the basis for inference in the model needs to be both biologically plausible as well as computationally tractable.
\label{sec:headings}

\subsection{Building Blocks}
This model assumes three key biological capacities or building blocks: the ability to detect coincident events (stimuli) and generate and modify\cite{aljadeff2019cortical,grant2017biologically} neuronal connections on that basis in a pseudo-Hebbian manner\cite{hebb2005organization,brown1994hebbian}; the ability to modulate connectivity (attenuation or amplification) between neurons (nodes) at their synapses (edges); and the availability of reward/penalty functions\cite{singh2009rewards} (such as those provided by dopamine\cite{bromberg2010dopamine}) as the fundamental building blocks of the model. 

The coincidence detection function takes "external" inputs from an environment graph and generates a "coincident" graph which constitutes the internal representation of the agent. The attributes of the graph (dynamically changing node and edge values) provide the underlying geometry and metric used by the system in its computations. We call this graph the Coincident Graph.

As the agent is exposed to new external stimuli, additions of nodes and edges and the modulation of these edges in the coincident graph serve to provide the mechanism for learning and memory, as we will see below.

Finally, reward and penalty functions that are ‘pre-wired’ into the coincident graph (which may still be modulated) are fundamentally responsible for generating the signal gradient used by the agent for memory, inference and behavior. Rewards are created by increasing/amplifying the conductance at edges, while penalties are established by reducing those conductances.

The theoretical proposition is that a minimal set of information about coincident inputs\cite{williams2022suspicious} can generate an information landscape that when coupled with external inputs and the reward/penalty structure, establishes a gradient that allows the system a bias or choice towards one node or set of nodes or another in a graph\cite{zhang2023endotaxis}. The output of the system includes the distribution of node values, as well as the updated state of the dynamical coincident graph, which changes at every time step.

As coincident stimuli are detected, this connection information is  created\cite{vardalaki2022filopodia}and retained in a matrix which is the adjacency matrix of the coincident graph or internal representation of the environment held by the model. These connections are then modulated as new stimuli are experienced, providing the plasticity required for the system to learn. As the inputs and outputs are processed, these intermediate values are transient and are \textit{not retained} in the long term by the system; memory functions are instantiated only by passing new inputs through the coincident graph and \textit{re-generating a memory construct from the interaction of current stimuli and the evolving state of the coincident matrix}. We will show that this mechanism not only gives the model a very compact and efficient way of retaining past experience, but also gives the system a flexible mechanism to generate new, compositional output structures as new inputs are connected in the coincident graph.

\subsection{Cellular Automata as Computation Model}

The selection of a cellular automaton-like computational model\cite{wolfram1984cellular} provides a concise way of describing the evolving state of the graph. As external inputs change node states, the cell update rules in the model propagate values to neighboring nodes and then on to their respective neighbors, in a recursive and localized manner. After a set number of recursions, the final node states are the “output” of the computation, as in most types of cellular automata\cite{neumann1966theory}.

The specific type of CA used in this model has certain unusual characteristics: firstly, the underlying graph is structurally dynamic\cite{alonso2007structurally,ilachinski2009structurally}, meaning that the network is not fixed, but itself also has an update function based on the type of external inputs received; secondly, the node and edge values are continuous\cite{kaneko1991coupled} rather than discrete; and finally, the computation goes through a limited set of recursions before stopping, unlike, for instance, certain types of Hopfield networks that seek an equilibrium state before converging or halting\cite{hopfield1982neural}.

Two different graph representations are used in the model: the environment graph or topology which defines the external stimulus/choice landscape available to the agent during training and subsequent interaction/behavior, and the internal graph representation generated by the algorithm and stored in the coincident graph matrix. The CGL algorithm generates a signal gradient in the internal representation as it traverses the environment graph, which in turn determines the subsequent state and choices of the agent. The system is entirely mechanistic in its functioning, though as we’ll see there are key probabilistic elements that come into play both in the primary cell update function as well as in sparse-reward or ambiguous environments.

\section{Methods}

The system takes coincident sensory inputs from the environment and generates an internal graph representation of those sequences of stimuli\cite{montague1994predictive}. As in most CAs, there is an inherent clock in the model, and the assumption is that coincident events can be specified as sensory events happening within a very short time interval. The graph representation serves as the underlying computational framework and dynamic memory storage at the same time, as is common in cellular automata. 

As sensory inputs are received, those values represent the initial state values of the corresponding cells or nodes of the automata and are nominally set to 1 when triggered. The cells then follow the cell update function $\phi$ and generate an output state of cells based on the underlying graph or lattice. This function may recurse a set number of times, where the resulting continuous values of the cell update function are then fed back into the same function.

The graph update function $\psi$ then takes the recent inputs and updates the underlying structure of the graph, modifying the edge weights, which are also nominally set to 1 when first connected.

The final state of the cell values and the topology of the coincident graph is the output of the computation and may be used by the agent as the basis for choice/action within the environment. 

\subsection{Cell Update function $\phi$} In the CGL model we specify that the \textit{central} node’s value affects its immediate neighborhood \cite{barral2016synaptic}. Our cell update function is equivalent to the transition function for a lazy random walk, and in the case of each separate cell neighborhood, also equivalent to the stationary distribution of a random walk. At its core, the propagation function is probabilistic, and we can think of the model dynamics as a lossy Markov Random Field.

Given an input vector $\mathbf{x}$, a graph represented by the square coincident matrix $C$ of dimension $m$, the identity matrix $I$ and C's degree matrix $D$, the cell update function can be written in matrix form in (Equation 1) below.

\begin{equation}
    \large\phi(\mathbf{x})=\frac{1}{2}\mathbf{x}(D^{-1}\times C + I)
\end{equation}

where $D^{-1}$ is the inverse of the diagonal degree matrix of C.

In the model, the edge values in the adjacency matrix $C$ are specified as \textit{conductance} rather than \textit{resistance}.

\subsection{Recursion function $\omega$}
As in all cellular automata, a key element is recursion. The main function above generates an output state $\phi(x)$, and that output is fed back into the function $r$ times. This is expressed in the recursion function $\omega$ below:

\begin{equation}
    \large\omega(\mathbf{x}_n)=\phi^r(\mathbf{x}_n)
\end{equation}

Where the superscript $r$ denotes the number of recursions of $\phi$. It should be noted that the recursion may also carry over a latent or remainder value $l$, expressed in (Equation 3) below. However, this latency value is set to zero in the simulations presented in this paper.

\begin{equation}
    \large\omega(\mathbf{x}_n)=\phi^r(\mathbf{x}_n)+l\phi(\mathbf{x}_{n-1})
\end{equation}

\subsection{Example of Cell Update and Recursion functions}

Below is an simple example of the functions $\phi$ and $\omega$ operating on an input set [1,0,0]. The coincident graph $C$ shows that Node 1 is connected to Node 2 and Node 2 is connected to Node 3. The first input value is passed through the function twice (Equations 4 and 5).

\begin{equation}
    \frac{1}{2}\times[1,0,0]\times
    \begin{pmatrix}
\begin{bmatrix}
1 & 0 & 0 \\
0 & .5 & 0 \\
0 & 0 & 1
\end{bmatrix}
\times
\begin{bmatrix}
0 & 1 & 0 \\
1 & 0 & 1 \\
0 & 1 & 0
\end{bmatrix}
+
\begin{bmatrix}
1 & 0 & 0 \\
0 & 1 & 0 \\
0 & 0 & 1
\end{bmatrix}
\end{pmatrix}
=[0.5,0.5,0]
\end{equation}
\begin{equation}
    \frac{1}{2}\times[.5,.5,0]\times
    \begin{pmatrix}
\begin{bmatrix}
1 & 0 & 0 \\
0 & .5 & 0 \\
0 & 0 & 1
\end{bmatrix}
\times
\begin{bmatrix}
0 & 1 & 0 \\
1 & 0 & 1 \\
0 & 1 & 0
\end{bmatrix}
+
\begin{bmatrix}
1 & 0 & 0 \\
0 & 1 & 0 \\
0 & 0 & 1
\end{bmatrix}
\end{pmatrix}
=[0.375,0.5,0.125]
\end{equation}

\subsection{Matrix Update function $\psi$}

In addition to the cell update and recursion functions above, the underlying coincident graph is also dynamic, and changes according to the external sensory inputs received by the system. Coincident nodes are connected when simultaneous external signals are received initially, and then de-inforced if detected again. This specific de-inforcement function has been observed as an inhibitory effect in multiple biological systems\cite{jacob2021prior,garner2022cortical}. The logical elements comprise:

\begin{itemize}
\item 
If two inputs are seen coincident for the first time, the edge value is set to one.
\item 
If two inputs are seen coincident once again, the current connection value is decreased by a factor $d$ (the factor is set to 0.5 in the simulations). This de-inforcement can repeat, down to a minimum value $b$.
\item 
If no simultaneous inputs are detected at that time window, nothing is changed.
\end{itemize}
We can express the matrix update rule as follows:

\begin{equation}
  \phi(C_{ij})=\begin{cases}
    x_ix_j & \text{if $C_{ij}=0$ and $i\neq 0$}.\\
    dC_{ij} & \text{if $C{ij}>b$ and $i\neq 0$ and $x_ix_j\neq 0$}.\\
    C_{ij} & \text{otherwise}.
  \end{cases}
\end{equation}

Where $d$ is the de-inforcement factor, and $b$ is the minimum “floor”  below which the edge value is no longer de-inforced.

The use of \textit{de-inforcement} rather than \textit{re-inforcement} in the function has implications for enabling exploratory behaviors for the agent. This will be addressed further in the Discussion section.

\subsection{Reward and Penalty functions}

The reward and penalty functions are represented as specialized nodes and/or edges within the coincident graph or matrix itself, and are treated as an intrinsic reward or penalty, in the sense that connections to those nodes are pre-wired into the matrix. The reward nodes/edges or penalty edges are connected to the relevant sensory nodes in the graph. Reward values are coded into the coincident graph $C$ as positive values above one, and penalties as values between zero and one in the experiments. 

\subsection{Memory Storage, Regeneration, and Expression}

In the CGL model, most of the inputs and outputs are discarded after being processed. Therefore, there is no separate \textit{location} of memory storage in the model, but rather, the memory function is generated by the interaction between the most recent input vector and the current state of the graph structure $C$.

One can think of the discrete distribution of the output vector as the \textit{current} experience of the agent regardless of the actual environmental input, which is the input vector. We can simply define the memory vector $\mathbf{m}$ as being the difference between the final output experience $\omega(\mathbf{x})$ and the actual current external sensory input vector $x$ (Equation 7).

\begin{equation}
   \large\mathbf{m}=\omega(\mathbf{x})-\mathbf{x}. 
\end{equation}

In this model, memory is a \textit{partial regeneration of a previous experience}, dependent on the interaction between current inputs and the state of the coincident graph. Note that we do not need to use the value of $\mathbf{m}$ directly in the model, but it is nonetheless interesting to note its definition in this way, as we can simply subtract the discrete distribution of the external input stimulus from the distribution of the output state of the automaton and define a concept of memory concisely.

By regenerating experiences in this way, the representation of past inputs is efficiently contained in the coincident graph connections and their modulations rather than stored in a separate memory location\cite{martin2000synaptic,roy2022brain,ryan2015engram,tonegawa2015memory}. We will see that this not only results in a highly compact representation, but also one that is flexible as new connections are made. This has interesting implications in the CGL system’s manner and ability to handle compositionality as well as context. 

\subsection{Experimental Design}

The experiments below are designed to show that an agent using the CGL functions described and specified above can use sensor inputs to generate the information space (coincident graph) based on its navigation of an environment, and once within range of a reward or penalty signal gradient, can make choices to move towards (or away from) those signals. The range of the gradient signal is directly related to the number of recursions that the cellular automata can carry out.

In all the experiments/simulations carried out, the external environment or topology is specified as a environment graph, which is then traversed by the agent during a single training run. \textit{The agent has no coordinate system or sense of absolute location within the environment}. The agent only perceives the node inputs that are directly adjacent to its node position in the graph, and exposure to those coincident inputs generates the coincident graph that is the agent’s internal representation of the environment, including both connection and conductance information. 

Reward and/or penalty nodes/circuits are established prior to agent training sessions, and are established within the coincidence matrix itself, and not as an input value. Note that the manner in which a reward circuit is activated is not by increasing the target node value directly, \textit{but by increasing the conductance on the edge between it and a specialized reward node}.

Behavior/navigation is modeled by simply moving the agent to the highest value adjacent node (including the self-node) from the active node's neighborhood. The path therefore is not pre-computed, but is dynamic and emergent based on the changing state of the coincident graph. If more than one neighboring node have equal high values, then one of them is chosen at random. it should be noted that the path of an agent that is out of range of a returning reward signal is still being influenced by de-inforcement, pushing the agent in an exploratory fashion away from past node locations.

Experiments are carried out in a variety of different environment graph environments, from highly ordered or regular lattices to increasingly randomized Watts-Strogatz small world networks with varying levels of connectivity and diameters.

In all experiments, the initial training run consists of a single complete pass through the environment. Note that all settings, parameters such as range, de-inforcement and rewards \textit{are identical for the training period as it is during the subsequent trial}. In other words, the agent is operating in the same state in both cases, save for navigation. After the single training run, the agent is placed in arbitrary or randomized initial locations and then responds to the signal gradient that is changing at every step.

A successful run is defined as reaching the reward location within a maximum number of steps equal to the number of nodes in the graph and remaining on it for at least 3 steps after reaching it.

In addition to individual experiments exploring specific reward/penalty behaviors, we run a set of randomized path trials on lattice graphs as well as three types of Watts-Strogatz graphs, from highly ordered to fully randomized. The random trials each test a sample of 400 random pairs of 100 start/target nodes analyzing success rates and successful path lengths and distributions for each trial.

\section{Results}

\subsection{Gradients, Rewards, and De-inforcement Generate Basis for Inference and Choice}

To observe the basic behavior of the CGL function in a simple setting, the agent was first trained on a simple, 25-node graph connected linearly (like a line of train carriages). After the training run, we find that with no de-inforcement or reward/penalty activation, placing the agent in the middle of the graph at node 13 with a recursive range of 5 generates a symmetrical gradient of values around the location in the general shape of a binomial distribution (Fig.3a). Connecting a wired reward node 26 with a target node at 16 result in a skewing of the gradient towards the reward location (Fig.3b). When de-inforcement at 0.5 was introduced (Fig.3c), the highest value accessible node was 14, in the direction of the reward location. We believe the combination of these fundamental interactions of the CGL function to be the basic inference mechanism that allows the model to operate successfully.

\begin{figure}[h!]
    \centering
    \includegraphics[width=0.75\linewidth]{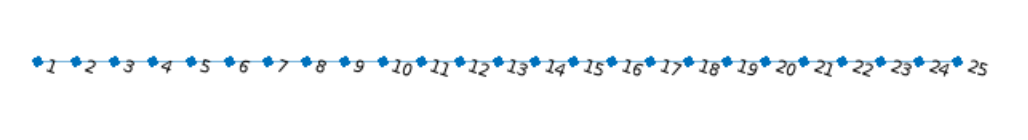}
    \caption{Environment graph or external topology.}

    \label{fig:enter-label}
\end{figure}
\begin{figure}
    \centering
    \includegraphics[width=0.4\linewidth]{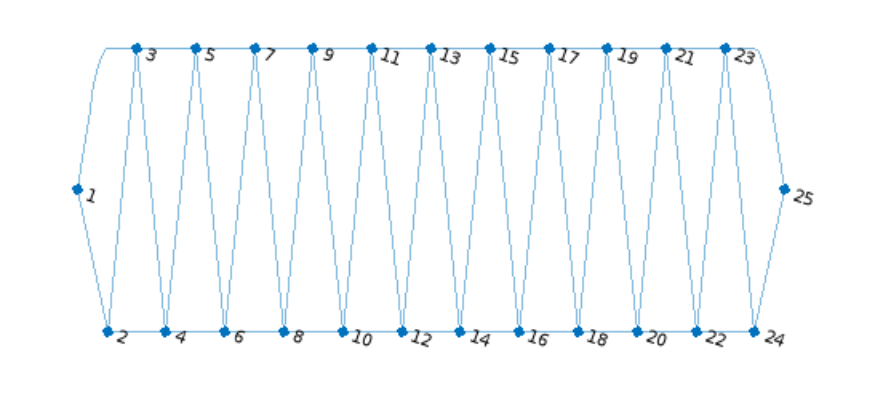}
    \caption{Coincident graph, or internal representation, after single training run.}
    \label{fig:enter-label}
\end{figure}

\begin{figure}[h!]
    \centering
    \includegraphics[width=.85\linewidth]{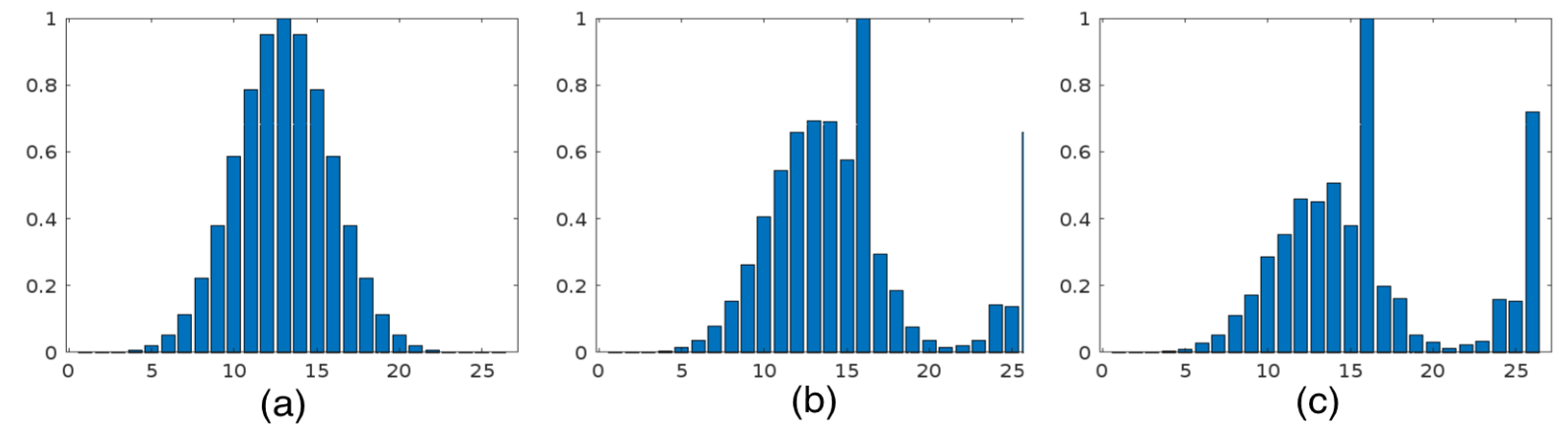}
    \caption{Node values exhibit a gradient across a simple graph. In Fig.3 an input at location 13 generates a set of values across a range of nearby nodes. In Fig 3a, the recursion creates a symmetrical gradient.  In  Fig 3b, a reward node is connected to node 16, In Fig 3c, the de-inforcement function is activated, providing the impetus for the agent to choose the highest visible value (node 14). Note that node values are normalized so max = 1.}
    \label{fig:enter-label}
\end{figure}

\subsection{Reward and Penalty Navigation in Lattice Grids}
Testing the model in an 8x8, 64-node lattice graph approximating a reinforcement learning gridworld task showed that the agent could successfully navigate to the reward-connected node after a single training run through the environment graph (Fig 4a). Note that in this series of experiments, a reward node 65 is connected to the target node at 64 (though node 65 is not shown in Figs. 4b and c it is there in all three cases). Penalties are coded as edges with reduced conductance. By adding penalty edges to a node in the middle of the path taken by the agent when the penalty was absent, we observe its adjustments as penalty edges are added (Fig 4b,c). More penalty edges resulted in the agent avoiding the penalty-connected node by a larger margin.

\begin{figure}
    \centering
    \includegraphics[width=0.85\linewidth]{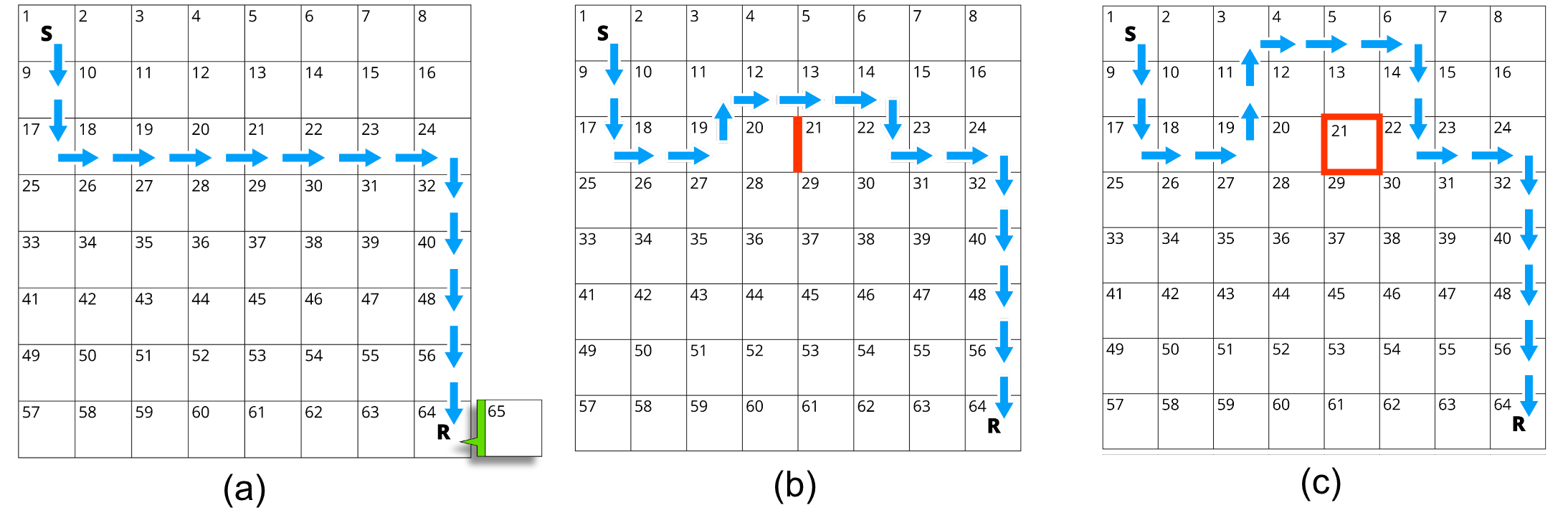}
    \caption{Agent behavior with reward node 65 connected to node 64 (labeled R) in Fig.4a and two configurations of penalty edges at node 21 in Figs.4b and 4c. Start node is node 1, labeled S.}
    \label{fig:enter-label}
\end{figure}

\subsection{More Complex Penalty Navigation}

We observe the behavior of the agent when trained in a more complex penalty environment:  a larger, 10x10, 100-node lattice graph with a varying set of penalty edges being added to its previous paths (Fig.5). As the agent avoids a penalty, it does not necessarily return to a previous successful path --- note that the agent is not \textit{pre-computing} routes, but assessing the output state one step at a time.

\begin{figure}[H]
    \centering
    \includegraphics[width=0.75\linewidth]{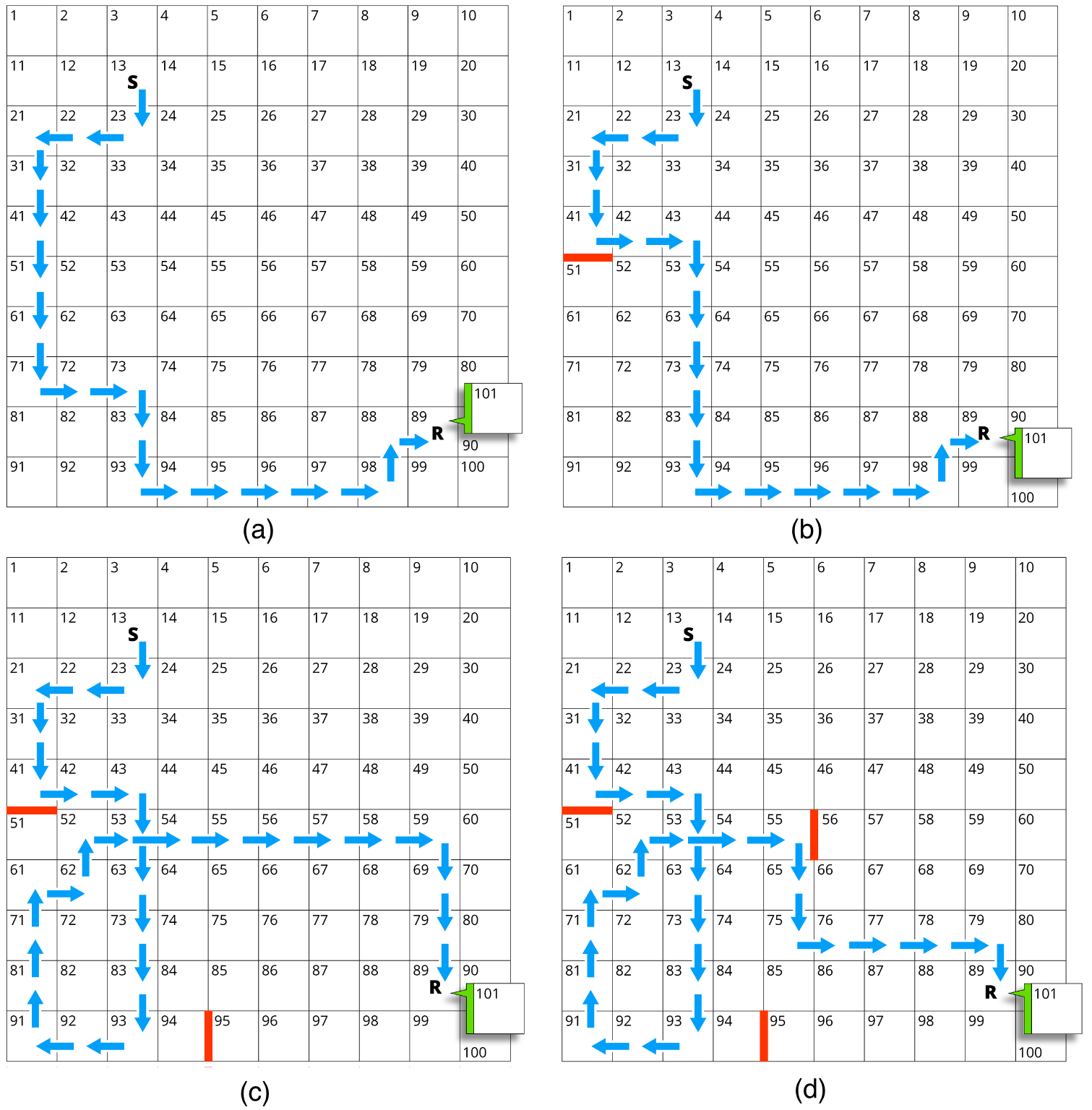}
    \caption{Navigating a more complex penalty environment. Fig 5a shows the path with no penalty nodes. Figs 5b, 5c and 5d show the agent responding to additional penalty edges.}
    \label{fig:enter-label}
\end{figure}

\subsection{Context and Compositionality with Multiple Rewards}

To demonstrate simple examples of compositionality and context-based decision-making(Fig.6), we train and agent with two rewards simultaneously, and then place the agent in different start locations (or more generally speaking, \textit{contexts} or initial conditions.

\begin{figure}
    \centering
    \includegraphics[width=0.74\linewidth]{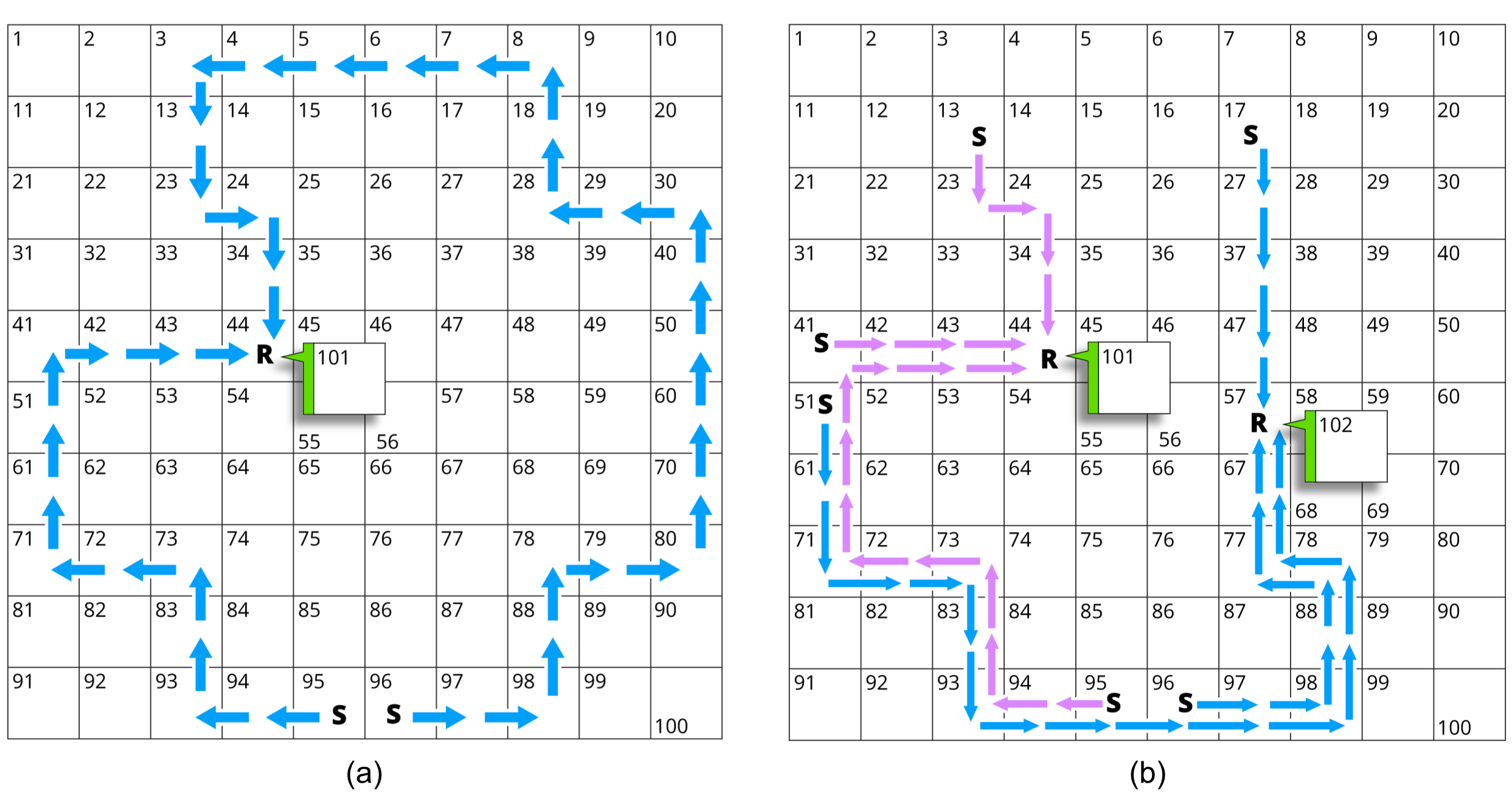}
    \caption{A simple demonstration of context-based (input vector) task-switching. Two rewards R1 and R2 at nodes 44 and 57 are trained at once, with equal (+3) weights. In Fig.6a two agents are trained on a single reward but started at different locations. In Fig.6b six agents are all trained on the same two rewards and started at different locations.}
    \label{fig:enter-label}
\end{figure}

\subsection{Observing the Gradient in Two Dimensions}

To better visualize the behavior of the signal gradient in the lattice-like environment tested in (Fig.4a) above, we generated a heatmap of all the the node values at each step of the agent run (Fig.7). Although the mechanism is the same in non grid-like graphs, it is easier to visualize the gradient and underlying dynamic in a 2-D setting.

\begin{figure}
   \centering
    \includegraphics[width=0.8\linewidth]{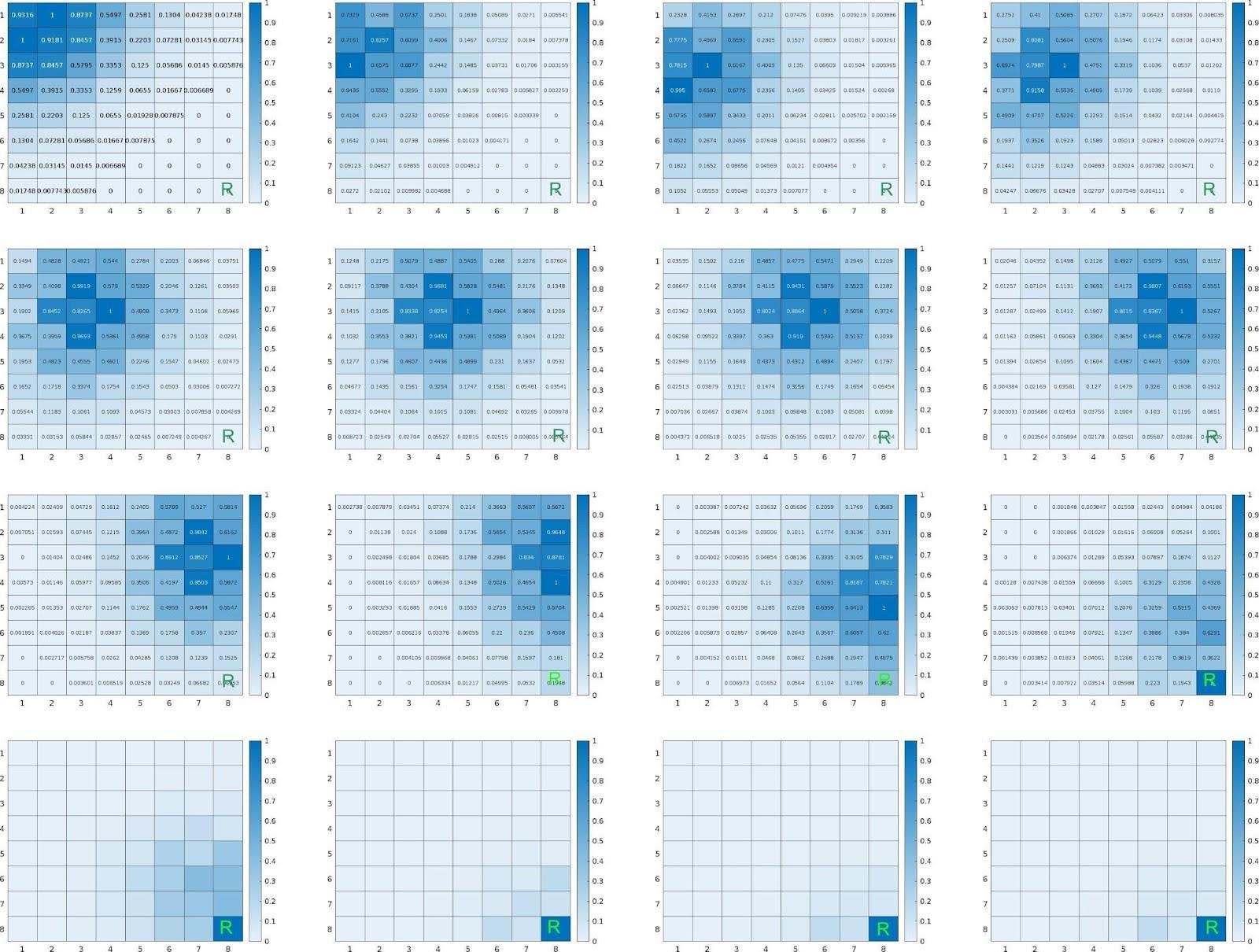}
    \caption{Heatmap of node values at each step as the agent moves from start location node 1 to target node 64 in the Experiment shown in Fig.4a}
    \label{fig:enter-label}
\end{figure}

\subsection{Randomized Experiments}

We have so far shown experiments on fairly ordered, lattice-like graph structures, which make it easier to observe and visualize the dynamics of the CGL system. However, in order to simulate somewhat more realistic environments, we simulate an broader range of environment graphs, from ordered ring lattices to more randomized Watts-Strogatz networks. Each of the experiments used a single training run over a 100-node graph over which we tested 400 random pairs of start/target nodes (total population of permutations: 9900), allowing a maximum of 100 total steps to find and stay on the target node for a minimum of three steps. We show success rates in the range of 92\% to 100\% for all 4 trials, with median number of steps to target in the range of 8 to 19. The results of the randomized trials are shown in (Figs.8-11) below. We note that reward and recursion parameters were changed between trials, as indicated in the figures. We note that the average diameters of the more ordered graphs are much larger than the more random ones, possible explaining the relationship to the number of recursions.

\begin{center}
\textbf{100 Node Lattice Graph}
\end{center}

\begin{table}[h]
\begin{center}
    \begin{tabular}{|c|c|} \hline 
         Success Rate& 98.5 \%\\ \hline 
         Median Path Length& 19\\ \hline 
         Mean Path Length& 20.45\\ \hline 
         Reward Level& 3\\ \hline 
         Recursions& 6\\ \hline
    \end{tabular}   
    \end{center}
\end{table}

\begin{center}
    \includegraphics[width=0.8\linewidth]{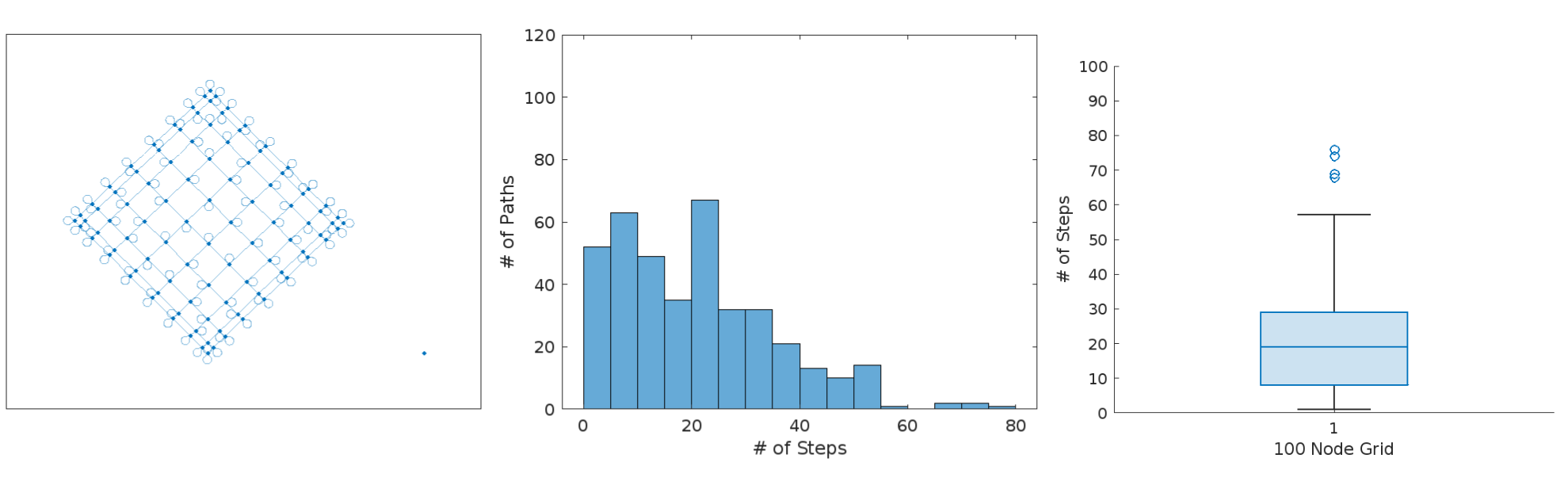}
    \captionof{figure}{100-node lattice graph, 400 random start/end points, showing histogram and box plot of path lengths.}
    \label{fig:enter-label}
\end{center}

\begin{center}
\textbf{0 \% Randomized Watts-Strogatz Graph}
 \end{center}

\begin{table}[h]
\begin{center}
    \begin{tabular}{|c|c|} \hline 
         Success Rate& 100 \%\\ \hline 
         Median Path Length& 8\\ \hline 
         Mean Path Length& 7.92\\ \hline 
         Reward Level& 5\\ \hline 
         Recursions& 6\\ \hline
    \end{tabular}  
    \end{center}
\end{table}
\begin{center}
    \includegraphics[width=0.8\linewidth]{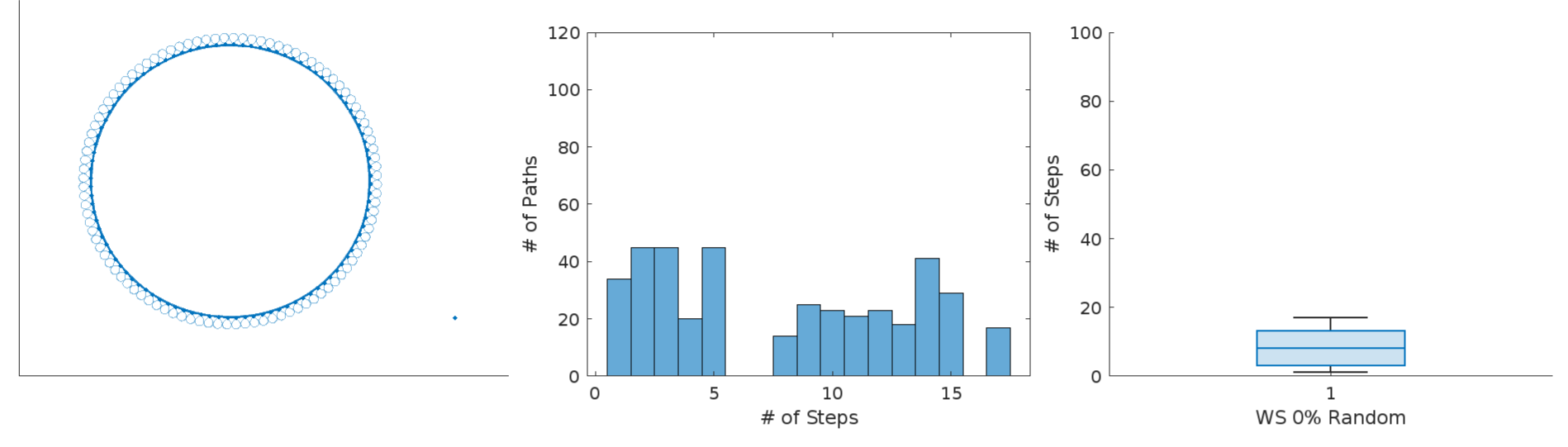}
    \captionof{figure}{ 100-node, 0\% randomized Watts-Strogatz graph, 400 random start/end points, showing histogram and box plot of path lengths.}
    \label{fig:enter-label}
\end{center}

 \begin{center}
\textbf{50 \% Randomized Watts-Strogatz Graph}
 \end{center}

\begin{table}[h]
\begin{center}
    \begin{tabular}{|c|c|} \hline 
         Success Rate& 94.25 \%\\ \hline 
         Median Path Length& 9\\ \hline 
         Mean Path Length& 12.9\\ \hline 
         Reward Level& 3\\ \hline 
         Recursions& 3\\ \hline
    \end{tabular} 
    \end{center}
\end{table}

\begin{center}
    \includegraphics[width=0.8\linewidth]{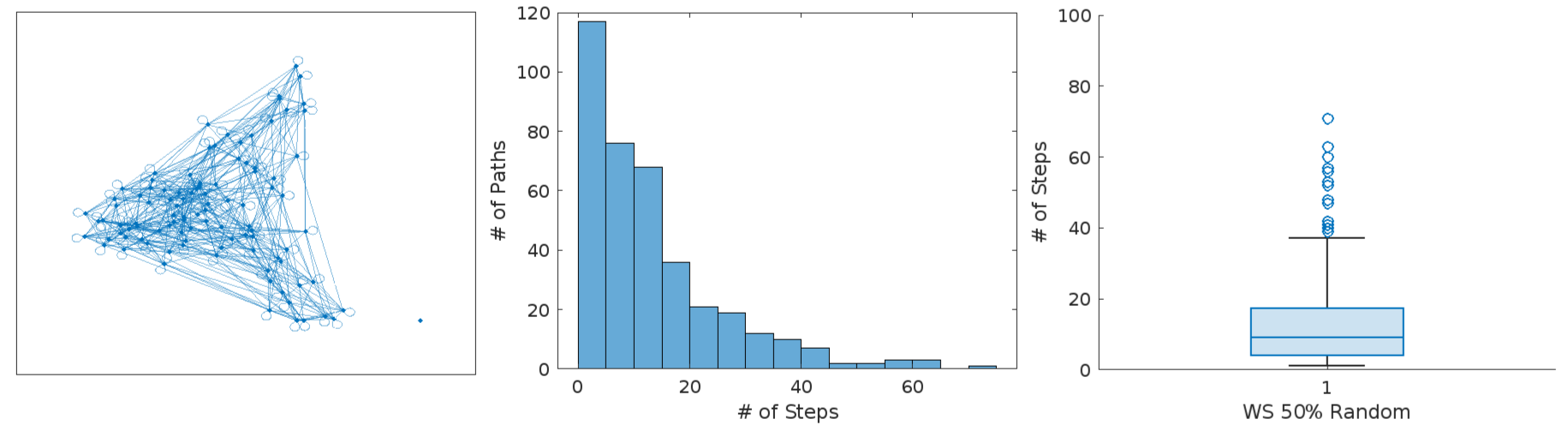}
    \captionof{figure}{100-node, 50\% randomized Watts-Strogatz graph, 400 random start/end points, showing histogram and box plot of path lengths.}
    \label{fig:enter-label}
\end{center}

\begin{center}
\textbf{100 \% Randomized Watts-Strogatz Graph}
\end{center}

\begin{table}[h]
\begin{center}
    \begin{tabular}{|c|c|} \hline 
         Success Rate& 92 \%\\ \hline 
         Median Path Length& 8\\ \hline 
         Mean Path Length& 13.43\\ \hline 
         Reward Level& 3\\ \hline 
         Recursions& 3\\ \hline
    \end{tabular} 
    \end{center}
\end{table}

\begin{center}
    \includegraphics[width=0.8\linewidth]{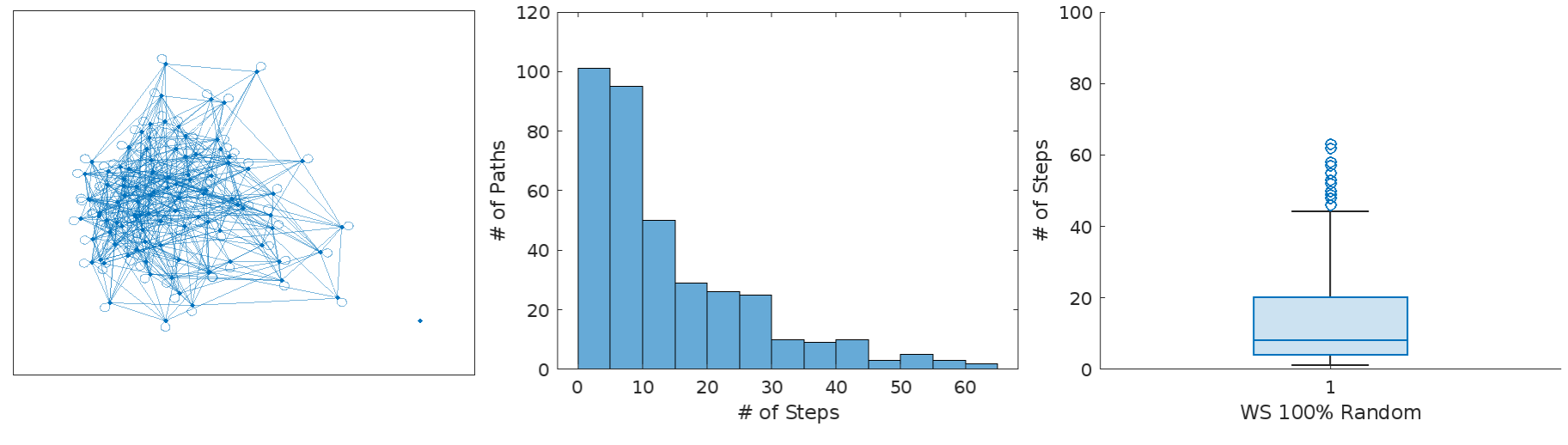}
    \captionof{figure}{100-node, 100 \% randomized Watts-Strogatz graph, 400 random start/end points, showing success rate within 100 steps and histogram of path lengths.}
    \label{fig:enter-label}
\end{center}

\section{Summary and Discussion}

We have shown in the experiments above that a simple recursive function based on coincident event detection, accompanied by a function that modulates edges in a graph environment with reward and penalty signals, is sufficient to generate the basis for elementary learning behaviors. We see that the choice of a structurally dynamic cellular automaton with limited time-steps is both efficient and generalizeable as a computational model. The experimental results show the capacity of the system to navigate a range of reward and penalty problems, and high success rates (between 92\% to 100\%) in finding solutions in randomized trials in a variety of small graph environments with different graph topologies.

We have shown in all the simulations that these functions are achievable with only a single training run through the environment, in contrast to comparable machine learning techniques, although it's difficult to compare the methods directly at this point.

We show that the interaction of input signals in the graph, propagating through a dynamically changing configuration and in the presence of amplification/attenuation from reward/penalty factors, generates a  gradient field that can be used as the basis for inference in the model, and the creation of dynamic, emergent navigational paths without the requirement of a specific, pre-computed or stored path.

We suggest that coincidence detection might be a plausible and fundamental sensory input that can be the basis for the information space and related distance metric used in biological computation\cite{threlfell2012striatal,montague1994predictive}.

We note that a parameter within the model that needed to be changed when testing more highly randomized Watts-Strogatz type graphs\cite{muldoon2016small} was the recursive range $r$, where we reduced it from 6 to 3 compared to the highly ordered ring graph or lattice grid. We believe that this is directly linked to the average diameters and connectivity of the graphs in question, which are smaller in the less ordered graphs, and where too much recursion generates noise in the gradient and obscures the signal from the reward and penalty locations.

We show that the coincident information represented in the graph is capable of regenerating a memory function without long-term storage of any specific inputs or outputs, resulting in a very compact and flexible representation of the past within the topology and conductances of the coincident graph itself. We suggest that this may have implications in understanding and building robust representational systems, biological or artificial, and may also provide a model for knowledge/memory transfer between autonomous agents in machine learning environments.

Although it may seem counter-intuitive that the primary axiomatic mechanism for plasticity is modeled as de-inforcement rather than re-inforcement, we note that it has certain advantages: from an energy efficiency point of view, reducing signal is cheaper than amplifying it, when it is really the relative values that generate the gradient of interest; it also provides a simple and elegant way of incorporating an exploration or \textit{novelty-seeking} function into the agent’s behavior when the gradient is equal (or zero) in all directions - in other words, it may better expressed as \textit{boredom avoidance}; we conjecture that the de-inforcement function may also act as a way of correcting/normalizing the internal graph representation since the graph training may evolve in non-systematic ways and coincident events are revisited by "accident"; and finally, recent evidence points to bio-neurological mechanisms that support this type of de-inforcement or inhibition function\cite{jacob2021prior,garner2022cortical}. We conjecture that during “rest” states of the agent with limited or no external stimuli, this de-inforcement needs to be lifted back up to some threshold state for the agent to continue to compute/operate properly, or the gradient will flatten/degrade to a higher entropy state. Much additional analysis of this mechanism is required.

In the experiments with multiple rewards, we show that the input vector represents a specific context that allows the agent to navigate to a different reward depending on that input. This may be a key attribute that could enable seamless task-switching, as well as a way to add new rewards, penalties, and experiential connections to the system without overwriting existing configurations in the coincidence matrix.. In other words, the model allows for a flexible way of generating compositionality or combinatorial generalization of output based a set of context-dependent connections in the matrix. This has some interesting implications in the AGI area of “lifelong learning machines”\cite{kudithipudi2022biological,van2020brain} and numerous related challenges being addressed in the AGI field in general, including but not limited to the problem of \textit{catastrophic forgetting}. It also has interesting implications as a model for the understanding of language, labels, and semantic meaning in the cognitive sciences and linguistics.

In the development of the CGL method, the intention was to use a minimum set of components that would force the model to be computationally and energetically efficient. Though our test environments in this paper are admittedly small, the capacity of the model to train on environments in a single pass and the low computational requirements of the main functions compare favorably with the techniques used in many machine learning algorithms. Furthermore, we conjecture that the statistical attributes of the larger and more randomized graphs that are likely to occur in real-world settings lend themselves well to the CGL approach where the recursive range of the system does not have to be extreme to be effective.

We suggest that the recursive range $r$ can be thought of as the abstraction capacity of the system, and related to a measure of the logical depth of the computation involved. However, as the CA is inherently parallel in conception, the computational complexity is likely driven by a combination of this logical depth in combination with the final “width” or number of connected and active nodes in the output state. More analysis of this topic is required.

Finally, in considering computational energy efficiency, we have imagined the core function as an potentially reversible continuous logic gate with some resemblance to memristor hardware\cite{sah2014brains}. It has equal numbers of outputs as inputs, and as pointed out above, flexible memory storage to avoid having to erase it. We suspect that the core function may be logically reversible, but more analysis of this topic is also required\cite{toffoli1987cellular}.

\section{Acknowledgements}

Thanks to Michael Sipser, for helping simplify some of the math, Laura Selfors, for advice on statistics, and Taru Muranen, for suggesting I come up with a hypothesis in the first place.

\bibliographystyle{apalike}  
\bibliography{references} 

\end{document}